\documentclass[conference]{IEEEtran}
\IEEEoverridecommandlockouts
\usepackage{cite}
\usepackage{times}
\usepackage{url}
\usepackage{amsmath,amssymb,amsfonts}
\usepackage{graphicx}
\usepackage{textcomp}
\usepackage[dvipsnames]{xcolor}
\usepackage{algorithm}
\usepackage{algpseudocode}
\usepackage{hyperref}
\usepackage{subcaption}

\newcommand*{\img}[1]{%
    \raisebox{-.3\baselineskip}{%
        \includegraphics[
        height=\baselineskip,
        width=\baselineskip,
        keepaspectratio,
        ]{#1}%
    }%
}

\usepackage{tikz}
\usetikzlibrary{shapes.geometric, arrows, positioning}
\tikzstyle{arrow} = [thick,->,>=stealth]
\tikzstyle{process} = [rectangle, font=\footnotesize, minimum width=1.6cm, minimum height=1.2cm, text centered, text width=1.6cm, draw=black, fill=gray!10]
\tikzstyle{token} = [ellipse, font=\footnotesize, minimum width=2cm, minimum height=0.7cm, text centered, text width=1.5cm, draw=black]
\usepackage{hhline}
\usepackage{colortbl}
\colorlet{Green}{LimeGreen!25!}
\colorlet{Blue}{RoyalBlue!25!}
\colorlet{Red}{BrickRed!25!}

\newcolumntype{x}{>{\columncolor{Green}}c}
\newcolumntype{y}{>{\columncolor{Blue}}c}
\newcolumntype{z}{>{\columncolor{Red}}c}

\def\BibTeX{{\rm B\kern-.05em{\sc i\kern-.025em b}\kern-.08em
    T\kern-.1667em\lower.7ex\hbox{E}\kern-.125emX}}
\begin{document}

\title{Graph-of-Tweets: A Graph Merging Approach to Sub-event Identification}

\author{\IEEEauthorblockN{Xiaonan Jing}
\IEEEauthorblockA{\textit{Computer and Information Technology} \\
\textit{Purdue University}\\
West Lafayette, IN, USA \\
jing@purdue.edu}
\and
\IEEEauthorblockN{Julia Taylor Rayz}
\IEEEauthorblockA{\textit{Computer and Information Technology} \\
\textit{Purdue University}\\
West Lafayette, IN, USA \\
jtaylor1@purdue.edu}
}

\maketitle

\begin{abstract}
Graph structures are powerful tools for modeling the relationships between textual elements. Graph-of-Words (GoW) has been adopted in many Natural Language tasks to encode the association between terms. However, GoW provides few document-level relationships in cases when the connections between documents are also essential. For identifying sub-events on social media like Twitter, features from both word- and document-level can be useful as they supply different information of the event. We propose a hybrid Graph-of-Tweets (GoT) model which combines the word- and document-level structures for modeling Tweets. To compress large amount of raw data, we propose a graph merging method which utilizes FastText word embeddings to reduce the GoW. Furthermore, we present a novel method to construct GoT with the reduced GoW and a Mutual Information (MI) measure. Finally, we identify maximal cliques to extract popular sub-events. Our model showed promising results on condensing lexical-level information and capturing keywords of sub-events.
\end{abstract}


\section{Introduction}
\label{sec:intro}
With Twitter and other types of social networks being the mainstream platform of information sharing, an innumerable amount of textual data is generated every day. Social networks driven communication has made it easier to learn user interests and discover popular topics. An event on Twitter can be viewed as a collection of sub-events as users update new posts through time. Trending sub-events can provide information on group interests, which can assist with learning group behaviours. Previously, a Twitter event has been described as a collection of hashtags \cite{feng:2015,yang:2018}, a (set of) named entity \cite{mcminn:2015}, a Knowledge Graph (KG) triplet \cite{qin:2018}, or a tweet embedding \cite{dhingra:2016}. While these representations can illustrate the same Twitter event from various aspects, it can be argued that a KG triplet, which utilizes a graph structure, exhibits richer features than the other representations. In other words, the graph structure allows more semantic relationships between entities to be preserved. Besides KG, other NLP tasks such as word disambiguation \cite{pina:2016,bevilacqua:2020}, text classification \cite{skianis:2018,yao:2019}, summarization \cite{nayeem:2017,yasunaga:2017}, and event identification \cite{tonon:2017,fedoryszak:2019} have also widely adopted graph structures. A graph $G=(V, E)$ typically consists of a set of vertices $V$ and a set of edges $E$ which describes the relations between the vertices. The main benefit of a graph structure lies in its flexibility to model a variety of linguistic elements. Depending on the needs, "the graph itself can represent different entities, such as a sentence, a single document, multiple documents or even the entire document collection. Furthermore, the edges on the graphs can be directed or undirected, as well as associated with weights or not" \cite{vazirgiannis:2018}. Following this line of reasoning, we utilize a graph structure to combine both token- and tweet-level associations in modeling Twitter events. 

Graph-of-Words (GoW) is a common method inspired by the traditional Bag-of-Words (BoW) representation. Typically, the vertices in a GoW represent the BoW from a corpus. In addition, the edges encode the co-occurrence association (i.e. the co-occurrence frequency) between the words in BoW. Although the traditional GoW improved upon BoW to include word association information, it still fails to incorporate semantic information. One may argue that, as previously mentioned, using a KG can effectively incorporate both semantic information and corpus level associations into the graph. However, any pre-existing KG, such as WordNet \cite{miller:1995} and FreeBase \cite{bollacker:2008}, cannot guarantee an up-to-date lexicon/entity collection. Therefore, we propose a novel vocabulary rich graph structure to cope with the constantly changing real-time event modeling.

In this paper, we employ a graph structure to model tokens, tweets, and their relationships. To the best of our knowledge, this is the first work to represent document level graphs with token level graphs in tweet representation. Our main contributions are the developments of 1) a novel GoT; 2) an unsupervised graph merging algorithm to condense token-level information; 3) an adjusted mutual information (MI) measure for conceptualized tweet similarities.

\section{Related Work}
\label{sec:lit}
Various studies have adopted graph structures to assist with unsupervised modeling of the diverse entity relationships.

\textbf{Event Clustering.} 
Jin and Bai \cite{jin:2016} utilized a directed GoW for long documents clustering. Each document was converted to a graph with nodes, edges, edge weights representing word features, co-occurrence, and co-occurrence frequencies respectively. The similarity between documents was subsequently converted to the similarity between the maximum common sub-graphs. With the graph similarity as a metric, K-means clustering was applied to maximum common document graphs to generate the clusters. 
Jinarat et al. \cite{jinarat:2018} proposed a pretrained Word2Vec embedding \cite{word2vec:2013} based GoW edge removal approach for tweet clustering. They constructed an undirected graph with nodes being the unique words in all tweets and edges being the similarity between the words. Token and tweet clusters were created by removing edges below a certain similarity value. However, pretrained embeddings can be prone to rare words in tweets, where abbreviations and tags are a popular means for delivering information. 

\textbf{Event stream detection.} 
Meladianos et al. \cite{meladianos:2015} presented a similar graph of words approach in identifying sub-events of a World Cup match on Twitter. They improved edge weight metric by incorporating tweet length to global co-occurrence frequency. The sub-events were generated by selecting tweets which contains the top k-degenerate subgraphs. 
Another effort by Fedoryszak et al. \cite{fedoryszak:2019} considered an event stream as a cluster chain consisting of trending Twitter entities in time order. The clusters were treated as nodes and the similarities between them were labeled as edge weights. While a common issue in entity comparison may be raised for the lack of coverage limitation \cite{mcminn:2015}, Fedoryszak et al. \cite{fedoryszak:2019} were able to overcome this issue through the help of an internal Twitter KG. However, potential synonyms were not taken into account in the weights assignment.

\textbf{Summarization.} 
Parveen and Strube \cite{parveen:2014} proposed an entity and sentence based bipartite graph for multi-document summarization, which utilized a Hyperlink-Induced Topic Search algorithm \cite{hits:1999}. However, they only considered nouns in the graph and ignored other parts-of-speech from the documents. 
In another attempt, Nayeem and Chali \cite{nayeem:2017} adopted the TextRank algorithm \cite{mihalcea:2004} which employs a graph of sentences to rank similar sentence clusters. To improve from the crisp token matching metric used by the algorithm, Nayeem and Chali \cite{nayeem:2017} instead used the pretrained Word2Vec embeddings \cite{word2vec:2013} for sentence similarities. 

\textbf{Graph Construction.} 
Glavas et al. \cite{glavavs:2016} built a semantic relatedness graph for text segmentation where the nodes and edges denote sentences and relatedness of the sentences respectively. They showed that extracting maximal cliques was effective in exploiting structures of semantic relatedness graphs.
In an attempt of automated KG completion, Szubert and Steedman \cite{szubert:2019} proposed a  word embedding based graph merging method in improvements of AskNET \cite{harrington:2008}. Similar to our merging approach, tokens were merged incrementally into a pre-constructed KG based on word embedding similarities. The difference is that AskNET used a graph-level global ranking mechanism while our approach considers a neighbor-level local ranking for the tokens. Additionally, Szubert and Steedman limited their scope to only named entity related relations during the graph construction.

\textbf{Evaluations of Event Identification.} It should be noted that there is no benchmark dataset for event identification evaluation, as many event identification approaches are unsupervised and that the event dataset varies by the research of interests. Among the previously mentioned studies, Jin and Bai \cite{jin:2016} conducted their clustering on dataset with category labels which allowed their unsupervised approach to be evaluated with precision, recall, and F-scores. Meladianos et al. \cite{meladianos:2015} was able to generate labels through a sport highlight website ESPN as the sub-events contained in a soccer game exhibit simpler structure than most real-life events. Fedoryszak et al. \cite{fedoryszak:2019} created a evaluation subset from scratch with the help of the Twitter internal KG and manually examined their clustering results. Both of the event summarization approaches adopted the classic  metric ROUGE \cite{lin:2004} score as well as the bench-marking dataset DUC.

\section{Methodology}
\label{sec:method}
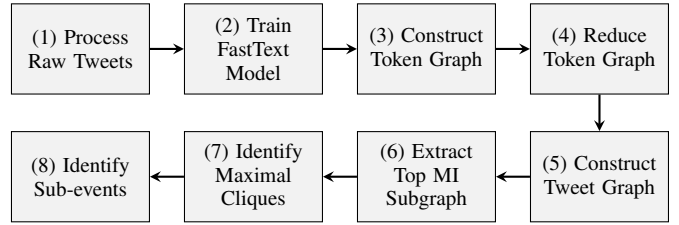
\begin{figure} [ht] 
    \centering 
    \begin{tikzpicture} 
        \node [process] (p1) {(1) Process Raw Tweets};
        \node [process, right of=p1, xshift=1.3cm] (p2) {(2) Train FastText Model};
        \node [process, right of=p2, xshift=1.3cm] (p3) {(3) Construct Token Graph};
        \node [process, right of=p3, xshift=1.3cm] (p4) {(4) Reduce Token Graph}; 
        \node [process, below of=p4, yshift=-0.7cm] (p5) {(5) Construct Tweet Graph};
        \node [process, below of=p3, yshift=-0.7cm] (p6) {(6) Extract Top MI Subgraph};
        \node [process, below of=p2, yshift=-0.7cm] (p7) {(7) Identify Maximal Cliques};
        \node [process, below of=p1, yshift=-0.7cm] (p8) {(8) Identify Sub-events};
        \draw [arrow] (p1) -- (p2);
        \draw [arrow] (p2) -- (p3);
        \draw [arrow] (p3) -- (p4);
        \draw [arrow] (p4) -- (p5);
        \draw [arrow] (p5) -- (p6);
        \draw [arrow] (p6) -- (p7);
        \draw [arrow] (p7) -- (p8);
    \end{tikzpicture} 
    \caption{8-Step Framework Used in This Paper}
    \label{fig:framework}
    \end{figure}

We propose the following $(1) - (8)$ steps for sub-event identification (Figure \ref{fig:framework}). The dataset used in this paper was collected from Twitter for a particular event (Section \ref{sec:dataset}). Step $(1)$ includes tokenization and lemmatization with Stanford CoreNLP and NLTK toolkit \footnote{Stanford CoreNLP ver. 3.9.2; NLTK ver. 3.4.5}, as well as removing stopwords, punctuations, links, and @ mentions. All processed texts are converted to lowercase. Steps (2) - (3) contribute to GoW construction (Section \ref{sec:gow_construction}). Step 4 performs our proposed GoW reduction method (Section \ref{sec:gow_reduction}). Step (5) elaborates on Graph-of-Tweets (GoT) construction using MI (Section \ref{sec:got}). Steps (6) - (8) finalize the subevents extraction from the GoT (Section \ref{sec:subgraph}). 

\subsection{Dataset}
\label{sec:dataset}
Following a recent trend on social media, we collected data on "COVID-19" from Twitter as our dataset. We define tweets containing the case-insensitive keywords \{"covid", "corona"\} as information related to the "COVID-19" event. We fetched 500,000 random tweets containing one of the keywords every day in the one month period from Feb 25 to Mar 25 and kept only original tweets as our dataset. More specifically, retweets and quoted tweets were filtered out during the data cleaning process. The statistics of the processed dataset used for FastText model training can be found in Table \ref{tab:dataset_all}. Besides FastText training, which requires a large corpus to achieve accurate vector representations, we focused on a single day of data, the Feb 25 subset, for the rest of our experiment in this paper.
    \begin{table}[ht] 
        \centering 
        \caption{Statistics of the dataset on COVID-19 from Feb 25 - Mar 25} 
        \begin{tabular}{|c|c|} \hline 
            number of tweets & 1,289,004 \\ \hline 
            number of unique tokens & 337,365 \\ \hline 
            number of tweets (Feb 25) & 38,508 \\ \hline 
            number of unique tokens (Feb 25) & 29,493 \\ \hline 
            average tokens per tweet & 12.85 \\ \hline 
            standard deviation on tokens per tweet & 6.89 \\ \hline 
        \end{tabular} 
        \label{tab:dataset_all} 
    \end{table}

\subsection{Graph-of-Words Construction} 
\label{sec:gow_construction}
We trained a word embedding model on the large 30-day dataset as our word-level similarity measure in GoW construction. Pretrained Word2Vec models \cite{word2vec:2013} have been applied previously to graphs as edge weights \cite{yasunaga:2017,jinarat:2018,skianis:2018}. Trained on Google News corpus, Word2Vec is powerful in finding context based word associations. Words appearing in similar contexts will receive a higher cosine similarity score based on the hypothesis that "a word is characterized by the company it keeps" \cite{firth:1957}. However, the pretrained Word2Vec model can only handle words within its vocabulary coverage. In other words, if any low frequency words are ignored during the training, they will not be captured in the model. In our case, the Twitter data contain many informal words, spelling mistakes, and new COVID-19 related words, which make pre-trained model not suitable for our task. On the other hand, FastText model \cite{fasttext:2017} uses character n-grams to enrich word representations when learning the word embeddings. Informal words such as "likeee" can be denoted as a combination of \{"like", "ikee", "keee"\} (4-gram), which its vector representation, after concatenating the subword vectors, will be closer to the intended word "like" given that both words were used in similar contexts. Thus, we employ the FastText word embeddings as the word-level similarity measure in our GoW construction. A skip-gram model with the Gensim implementation \footnote{Gensim ver. 3.8.1; hyperparameters (that differ from default): size: 300d, alpha: 0.05, min\_alpha=0.0005, epoch: 30} was trained on the large 30-day dataset. 

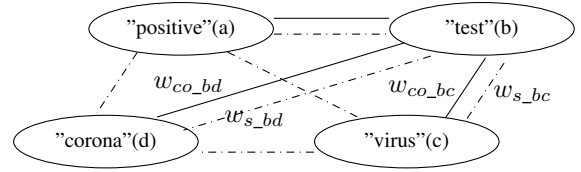
\begin{figure} [ht]
        \centering
        \begin{tikzpicture} 
            \node [token] (t1) {"positive"(a)};
            \node [token, right of=t1, xshift=3cm] (t2) {"test"(b)};
            \node [token, below of=t1, xshift=3cm, yshift=-0.5cm] (t3) {"virus"(c)};
            \node [token, below of=t1, xshift=-1cm, yshift=-0.5cm] (t4) {"corona"(d)};
            \draw [transform canvas={yshift=4pt}] (t1) -- (t2);
            \draw [transform canvas={xshift=8pt, yshift=-1pt}] (t2) -- node[left]{$w_{co\_bc}$} (t3);
            \draw [transform canvas={xshift=-5pt, yshift=2pt}](t2) -- node[above, xshift=-35pt, yshift=-8pt]{$w_{co\_bd}$} (t4);
            \draw [dash dot, transform canvas={yshift=-2pt}] (t1) -- (t2);
            \draw [dash dot] (t1) -- (t3);
            \draw [dash dot, transform canvas={xshift=15pt, yshift=-3pt}] (t2) -- node[right]{$w_{s\_bc}$} (t3);
            \draw [dash dot, transform canvas={xshift=-10pt, yshift=1pt}] (t1) -- (t4);
            \draw [dash dot, transform canvas={xshift=5pt, yshift=-3pt}] (t2) -- node[below, xshift=-20pt, yshift=-3pt]{$w_{s\_bd}$} (t4);
            \draw [dash dot, transform canvas={yshift=-4pt}] (t3) -- (t4);
        \end{tikzpicture} 
        \caption{An example GoW for the processed tweets \{"virus", "test", "positive"\} and \{"corona", "test"\}. Solid and dotted edges denote $E_{co}$ and $E_s$ respectively.}
        \label{fig:example_gow}
    \end{figure}

For the basic structure of the GoW, we adopt the approach from Meladianos et al. \cite{meladianos:2015} where the vertices $V$ represent the tokens and the edges $E_{co}$ represent the co-occurrence of two tokens in a tweet. For $k$ tweets in the corpus $T:\{t_1, t_2, ..., t_k\}$, the co-occurrence weight $w_{co}$ between the unique tokens $v_1$ and $v_2$ is computed as:
    \begin{equation} \label{eq:cooc_token}
        w_{co}= \sum^{k}_{i=1}{\frac{1}{n_i - 1}}
    \end{equation}
where $n_i (n_i > 1)$ denotes the number of unique tokens in tweet $t_i$ that contains both token $v_1$ and $v_2$. An edge $e_{co}$ is only drawn when the token pair co-occur at least once. In addition to the base graph, we add another set of edges $E_s$ denoting the cosine similarity $w_s$ between the token embeddings obtained from FastText. Figure \ref{fig:example_gow} illustrates an example GoW for two processed tweets. 

\subsection{Graph-of-Words Reduction}
\label{sec:gow_reduction}
A raw GoW constructed directly from the dataset often contains a large number of nodes. Many of these nodes carry repeating information, which increases the difficulties of the subsequent tasks. To condense the amount of nodes, we propose a two-phase node merging method to reduce the proposed GoW: 
    \begin{itemize} 
        \item Phase I: Linear reduction by node merging based on word occurrence frequency.
        \item Phase II: Semantic reduction by node merging based on token similarity. 
    \end{itemize} 

\textbf{Phase I.} The goal of this phase is to reduce the number of nodes in the raw graph in a fast and efficient manner. For tokens that occur in less than 5 tweets, we merge them to its top similar token node in the graph. This phase is performed on nodes in the order of least frequently appearing to most frequently appearing nodes.

\textbf{Phase II.} The goal of this phase is to combine frequent tokens in the graph. Algorithm \ref{alg:phase2} describes this process. 
    \begin{algorithm} [ht]
    \begin{algorithmic}
    \caption{Semantic Node Collapse}
    \label{alg:phase2}
    \For{node $v_i \in V:\{v_1, v_2, ..., v_m\}$} 
        \State degree $d_i = \sum^{m}_{1}{w_{co\_ij}}$
    \EndFor
    
    \State $V = \textit{\textbf{sort\_asc}}(V, key=d_i)$ 
    
    \For{$v_i \in V$}
        \State $sim\_neighbors = list()$ 
        \State $\textit{\textbf{neighbors}}(v_i) = B_i:\{v_{i1}, ..., v_{iz}\}$
        \State $\textit{\textbf{neighbor\_weights}}(v_i) = W_i:\{w_{co\_i1}, ..., w_{co\_iz}\}$ 
        \State $B_i = \textit{\textbf{sort\_asc}}(B_i, key=w_{co\_ij})$
        
        \For{$v_{ij} \in B_i$} 
            \State $sim\_token = \textit{\textbf{fasttext.most\_similar}}(v_{ij}, top\_n)$
            
            \If {any other neighbor $v_{ik} \in sim\_token$} 
                \State $sim\_neighbors.insert((v_{ik}, sim\_val))$ 
            \EndIf 
        
        \State $parent = \textit{\textbf{max}}(sim\_neighbors, key=sim\_val)$
        \State $\textit{\textbf{node\_merge}} (src=v_{ij}, dst=parent)$ 
    \EndFor
    \EndFor 
    
    \end{algorithmic}
    \end{algorithm}
For a node in the GoW, we merge its lower weighed neighbor into another neighbor if the top N similar token of the lower weighed neighbor contains another neighbor. It should be addressed that the ordering of the node and the direction of merging matters in this process. For token nodes in the GoW, we perform this phase on nodes in the order of lowest degree to highest degree; and for neighbors of the same node, we perform the phase on neighbors in the order of lowest weight to highest weight. When the top N similar token contains more than one other neighbors, we select the node with the highest similarity value as the parent node.

The \textbf{node merging} process consists of removing original nodes, reconnecting edges between neighbors, and migrating co-occurrence weights $w_{co}$ and merged tokens. Essentially, if a target node is merged into a parent node, the target node will be removed from the graph and the neighbors of the target node will be reconnected to the parent. It should be noted that when migrating neighbors and weights, we only consider the co-occurrence edge $E_s$ and only increment the weights $w_{co}$ into the edges of the parent node, while $w_s$ remains the same as determined by the leading token of the merged nodes. For a merged node with multiple tokens, we define the leading token as a single token that is carried by the original GoW. More precisely, suppose the target node $"corona"$ \footnote{in this paper, we will use the italic \textit{"leading token"} to represent a token node, and the normal "text" to represent plain text.} is to be merged into the parent node $"virus"$ in Figure \ref{fig:example_gow}. Since node $"corona"$ is only neighboring with node $"test"$, we add the weights together so that the new weight between node $"test"$ and $"virus"$ is $w_{co\_bc} = w_{co\_bd} + w_{co\_bc}$, and remove node $"corona"$ from the graph. Furthermore, for the new node $"virus"$ containing both tokens "virus" and "corona", the leading token is "virus", and the weight $w_{s\_bc}$ remains the same as the cosine similarity between the word vectors "virus" and "test".

\subsection{Graph-of-Tweets Construction with Normalized Mutual Information As Graph Edges}
\label{sec:got}
Similar to GoW, we introduce a novel GoT which maps tweets to nodes. In addition, to compare the shared information between the tweet nodes, we construct the edges with an adjusted MI metric. Each tweet is represented as a set of unique merged nodes obtained from the previous two-phase GoW reduction, and tweets with identical token node representation are treated as one node.  For instance, after merging token node $"corona"$ into $"virus"$ in Figure \ref{fig:example_gow}, a processed tweet \{"corona", "virus", "positive", "test"\} can be represented as a tweet node $t: \{"virus", "positive", "test"\}$ which contains the set of unique merged token nodes $"virus"$, $"positive"$, and $"test"$. 

Originally from Information Theory, MI is used to measure the information overlap between two random variables $X$ and $Y$. Pointwise MI (PMI) \cite{church:1990} in Equation \ref{eq:pmi} was introduced to computational linguistics to measure the associations between bigrams / n-grams. PMI uses unigram frequencies and co-occurrence frequencies to compute the marginal and joint probabilities respectively (Equation \ref{eq:marginal_bigram}), in which $W$ denotes the total number of words in a corpus where word $x$ and $y$ co-occur. 
    \begin{equation} \label{eq:pmi}
        i(x, y) = \log \frac{p(x, y)}{p(x) p(y)}
    \end{equation}
    \begin{equation} \label{eq:marginal_bigram}
        p(x) = \frac{f(x)}{W}, \hspace{3mm} p(y) = \frac{f(y)}{W}, \hspace{3mm} p(x, y) = \frac{f(x, y)}{W}
    \end{equation}
The drawbacks of PMI are: 1) low frequency word pairs tend to receive relatively higher scores; 2) PMI suffers from the absence of boundaries in comparison tasks \cite{bouma:2009}. In an extreme case when $x$ and $y$ are perfectly correlated, the marginal probabilities $p(x)$ and $p(y)$ will take the same value as the joint probability $p(x, y)$. In other words, when $x$ and $y$ only occur together, $p(x, y) = p(x) = p(y)$, which will result $i(x, y)$ in Equation \ref{eq:pmi} to take $- \log p(x, y)$. Therefore, with $W$ remaining the same, a lower $f(x, y)$ will result in higher PMI value. Additionally, it can be noted that PMI in Equation \ref{eq:pmi} suffers from the absence of boundaries when applied for comparisons between word pairs \cite{bouma:2009}. To mitigate the scoring bias and the comparison difficulty, a common solution is to normalize the PMI values with $- \log p(x, y)$ or a combination of $- \log p(x)$ and $- \log p(y)$ to the range $[-1, 1]$ to smooth the overall distribution. 

\begin{figure}[ht]
    \centering
    \includegraphics[width=6cm]{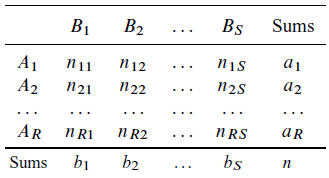}
    \caption{Contingency Table Between Clusterings A and B \cite{amelio:2017} }
    \label{fig:contigency_table}
    \end{figure}

Apart from measuring word associations, MI is also widely applied in clustering evaluations. In the case when the ground truth clusters are known, MI can be used to score the "similarity" between the predicted clusters and the labels. A contingency table (Figure \ref{fig:contigency_table}) is used to illustrate the number of overlapping elements between the predicted clusterings $A$ and ground truth clusterings $B$.
One disadvantage of using MI for clustering evaluation is the existence of selection bias, which Romano et al. \cite{romano:2014} described as "the tendency to choose inappropriate clustering solutions with more clusters or induced with fewer data points." However, normalization can effectively reduce the bias as well as adding an upper bound for easier comparison. Romano et al. \cite{romano:2014} proposed a variance and expectation based normalization method. Other popular normalization methods include using the joint entropy of $A$ and $B$ or some combinations of the entropies of $A$ and $B$ as the denominator \cite{amelio:2017}.

In our GoT case, since tweet nodes are represented by different sets of token nodes, we can treat the tweet nodes as clusters which allow repeating elements. Thus, the total number of elements is the number of token nodes obtained from the reduced GoW. Correspondingly, the intersection between two tweet nodes can be represented by the overlapping elements (token nodes). Following this line of logic, we define the normalized MI between two tweets in Algorithm \ref{alg:nmi}.

    \begin{algorithm} [ht]
    \caption{Normalized MI (NMI) Between Tweet Nodes}
    \begin{algorithmic} 
        \State Let $T: \{t_1, t_2, ..., t_k\}$ be the set of tweets in the GoT. 
        \State Let $V:\{v_1, v_2, ..., v_m\}$ be the $m$ token nodes in the reduced GoW. For tweets $t_i:\{v_{i1}, ..., v_{ix}\}$ and $t_j:\{v_{j1}, ..., v_{jy}\}$, the $MI$ is defined as: 
            \begin{center}  
                $MI(t_i, t_j) = \log \frac{p(t_i, t_j)}{p(t_i)p(t_j)}$ ,
            \end{center} \\
        where $p(t_i)$ and $p(t_j)$ are the probabilities of token nodes containing an individual tweet with respect of total number of token nodes $m$, with $p(t_i)=\frac{x}{m}$ and $p(t_j)=\frac{y}{m}$. 
        \State The joint probability $p(t_i, t_j)$ represents the intersection of token nodes between the two tweets, with $p(t_i, t_j)=\frac{count(t_i \cap  t_j)}{m}$.  
        
        \State To normalize $MI$ to the range $[-1, 1]$, we use a normalization denominator: \\
            \begin{center} 
                $norm = max[-\log p(t_i), -\log p(t_j)]$ , \\ 
                Thus, $NMI = \frac{MI}{norm}$ .
            \end{center}
        
        \State Note that when $p(t_i, t_j) = 0$ (indicating no intersection), NMI will take boundary value $-1$. 
    \end{algorithmic}
    \label{alg:nmi}
    \end{algorithm} 
    
It should be noted that as the fetching keywords \{"corona", "covid"\} appear in every tweet in the dataset, tokens containing these words are removed when the GoT is constructed. Consequently, two tweet nodes with only \textit{"corona"} or \textit{"covid"} in common will result in an NMI value of -1, while the outcomes of sub-event identification are not affected by removing the main event "COVID-19".

\subsection{Sub-event Extraction From Graph-of-Tweets}
\label{sec:subgraph}
We hypothesize that popular sub-events are included within a subgraph of GoT which are fully connected and highly similar in content. Following this assumption, we extract a GoT subgraph with only the tweet nodes included in the top\_n NMI values. Subsequently, we identify all maximal cliques of size greater than 2 from the subgraph for further analysis. As cliques obtained this way consist of only nodes with large NMI values, which indicates that the tweet nodes are highly similar, the clique can be represented by the token nodes included in the tweet nodes. Thus, we treat a clique as a set of all token nodes contained in the tweet nodes and that each clique represents a popular sub-event. 

\section{Results and Discussion}
\label{sec:results}
To summarize, the raw GoW consisted of 29,493 unique token nodes for tweets from the Feb 25 data division. The two-phase GoW reduction reduced the graph by 83.8\%, with 24,711 nodes merged and 4,782 token nodes left in the GoW. On the other hand, the raw GoT consisted of 31,383 unique tweet nodes. The extracted subgraph of the top 1000 MI values consisted of 1,259 tweet nodes. Finally, 83 maximal cliques were identified from the subgraph.

\subsection{Graph-of-Words Reduction} 
Phase I reduction merged 19,663 token nodes that appeared in less than 5 tweets, which is roughly 66.7\% of all token nodes contained in the raw GoW. This phase primarily contributes to reducing uncommon words within the corpus. It can be argued that rare words may provide non-trivial information. However, because our goal is to identify popular sub-events, the information loss brought by rare words does not heavily affect the results. Furthermore, word similarity provided by FastText can effectively map the rare terms to the more common words with a similar meaning. For instance, in our FastText model, the rare word "floridah" has its most similar word as the intended word "florida". Another fun fact is that the emoji "\img{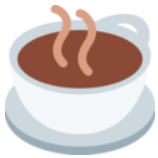}" was merged into the node \textit{"coffee"}. 

There were 5,048 token nodes merged during Phase II reduction. This phase mainly contributes to information compression within common tokens. By merging the neighbors that carry similar information, the same target node can appear in a less complex network without information loss. Table \ref{tab:merged_nodes} presents statistics of the resulting GoW. 
Table \ref{tab:example_merged_nodes} shows some example merged nodes from the reduced GoW \footnote{The full reduced Graph-of-Words (4,728 token nodes) can be found at: \href{https://github.com/lostkuma/GraphOfTweets}{https://github.com/lostkuma/GraphOfTweets}}. The largest node (248 tokens) is not shown here as it contains many expletives. The second largest node \textit{"lmaoooo"} (221 tokens) contains mostly informal terms like "omgggg" and emojis. 
    \begin{table}[ht]
        \centering
        \caption{Reduced GoW Statistics}
        \begin{tabular}{|c|c|} \hline 
            single / merged token nodes count & 2,119 / 2,663 \\ \hline 
            max / min node size & 248 / 2 \\ \hline 
            avg / std node size & 10.28 / 16.56 \\ \hline 
            max / min within merged node similarity & 0.9672 / 0.1337 \\ \hline 
            avg / std within merged node similarity & 0.4074 / 0.1544 \\ \hline 
        \end{tabular}
        \label{tab:merged_nodes}
    \end{table}
    
    \begin{table}[ht]
        \centering
        \caption{Some Example Merged Nodes After Reduction}
        \begin{tabular}{|p{1.8cm}|p{6cm}|} \hline 
            \textbf{leading token} & \textbf{merged tokens} \\ \hline 
            airline & frontier, iata, swoop, airliner, pilot, delta, piloting \\ \hline 
            cdc & wcdc, cdcS, cdcwho, cdcas, lrts, rts, lrt, \#lrt, cdcgov, cdcgovs, \#cdcgov, cdcemergencys, mmwr, \#mmwr, nih, cdcnih, \#nih, \#nihonbuyo, 514v \\ \hline
            expects & warns, \#warns, expected, expecting, expectin, brace, expect, therell \\ \hline 
        \end{tabular}
        \label{tab:example_merged_nodes}
    \end{table}

We identify roughly three patterns of merge, namely 1) words with same stem or synonym merge, 2) common bi-gram or fix-ed expression merge, and 3) words of topical related but semantically different merge during Phase II reduction. Table \ref{tab:example_merges_2} illustrates some example merges from source node to destination node, with the green , blue, and red columns correspond to type-1, type-2 and type-3 respectively. 
    \begin{table}[ht]
        \centering
        \caption{Some example merges in Phase II reduction}
        \setlength{\tabcolsep}{0.4em}
        \setlength{\arrayrulewidth}{0.1em}
        \begin{tabular}{|x|x||y|y||z|z|} \hline 
            \textbf{src node} & \textbf{dst node} & \textbf{src node} & \textbf{dst node} & \textbf{src node} & \textbf{dst node}\\ \hline  
            covid19 & covid & positive & test & buy & sell  \\ \hline 
            covid & coronavirus & department & health & always & never \\ \hline 
            \#iphone & \#apple & cruise & ship & masked & unmasked \\ \hline 
            complains & complaint & valley & silicon & men & woman \\ \hline 
            btc & bitcoin & long & term & latter & splatter \\ \hline 
            dead & death & conspiracy & theory & undo & mundo \\ \hline 
        \end{tabular}
        \label{tab:example_merges_2}
    \end{table}
Among type-1 merge (green), it can be seen that common abbreviations such as "btc" as "bitcoin" are also captured in the merging process. In type-2 merge (blue), the examples such as "test positive" and "health department" are frequent bi-grams in the context of our data; and other examples such as "silicon valley" and "long term" are fixed expressions. One general drawback of word embedding models is that instead of semantic similarity, words with frequent occurrence will be considered as very similar as noted in the distributional hypothesis \cite{firth:1957}. An improvement can be made by combining named entities, fixed expressions, and frequent bi-grams in the data processing stage so that a node can also represents a concept in the GoW. Finally, type-3 merge (red) is also suffers from the drawback of word embedding models. Word pairs like "buy" and "sell", "always" and "never", "masked" and "unmasked" are antonyms/reversives in meaning. However, these word pairs tend to be used in the same context so they are considered highly similar during the merging process. Word pairs like "latter" and "sp[latter]", "undo" and "m[undo]" (means "world" in Spanish) are subwords of each other. Recall that FastText model uses character n-grams in the training. The subword mechanism leads the rare words to be considered as similar to the common words which share a part with them. It should be noted that in Phase II, the merging is performed from the lower weighed neighbors to the higher weighed neighbors and from lower degree nodes to higher degree nodes. It is plausible that a common word like "undo" is a lower weighed neighbor as compared to the uncommon word "mundo" if the target token is also a rare word in Spanish and that "undo" does not co-occur frequently with the target token.

\subsection{Graph-of-Tweets and Sub-events Extraction} 
Using the reduced GoW, we constructed a GoT, where each tweet node was represented as a set of leading tokens from the token nodes. Of the original 38,508 tweets, 31,383 tweets were represented as unique tweet nodes. Take the tweet node \textit{tweet\_18736}, which repeated the most times with a frequency of 221 times, as an example. The original text of \textit{tweet\_18736} says "US CDC warns of disruption to everyday life' with spread of coronavirus \href{https://t.co/zO2IovjJlq}{https://t.co/zO2IovjJlq}", which is an event created by Twitter on Feb 25. After preprocessing, we obtained \{"u", "cdc", "warns", "disruption", "everyday", "life", "spread", "coronavirus"\}. Finally, the GoW leading token represented tweet node became \{\textit{"probably", "cdc", "expects", "destroy", "risking", "spreading", "coronavirus"}\}, with both "everyday" and "life" merged to \textit{"risking"}. One may notice that the word "US" is converted to "u" after preprocessing, which consecutively affected the FastText training, the GoW reduction and the GoT representation. This is due to a limitation from the WordNet lemmatizer that mapping the lower case "us" to "u". We would consider keeping the uncased texts and employ named entity recognition to identify cases like this to preserve the correct word in future work.

\begin{figure}[ht]
    \centering
    \includegraphics[width=0.48\textwidth]{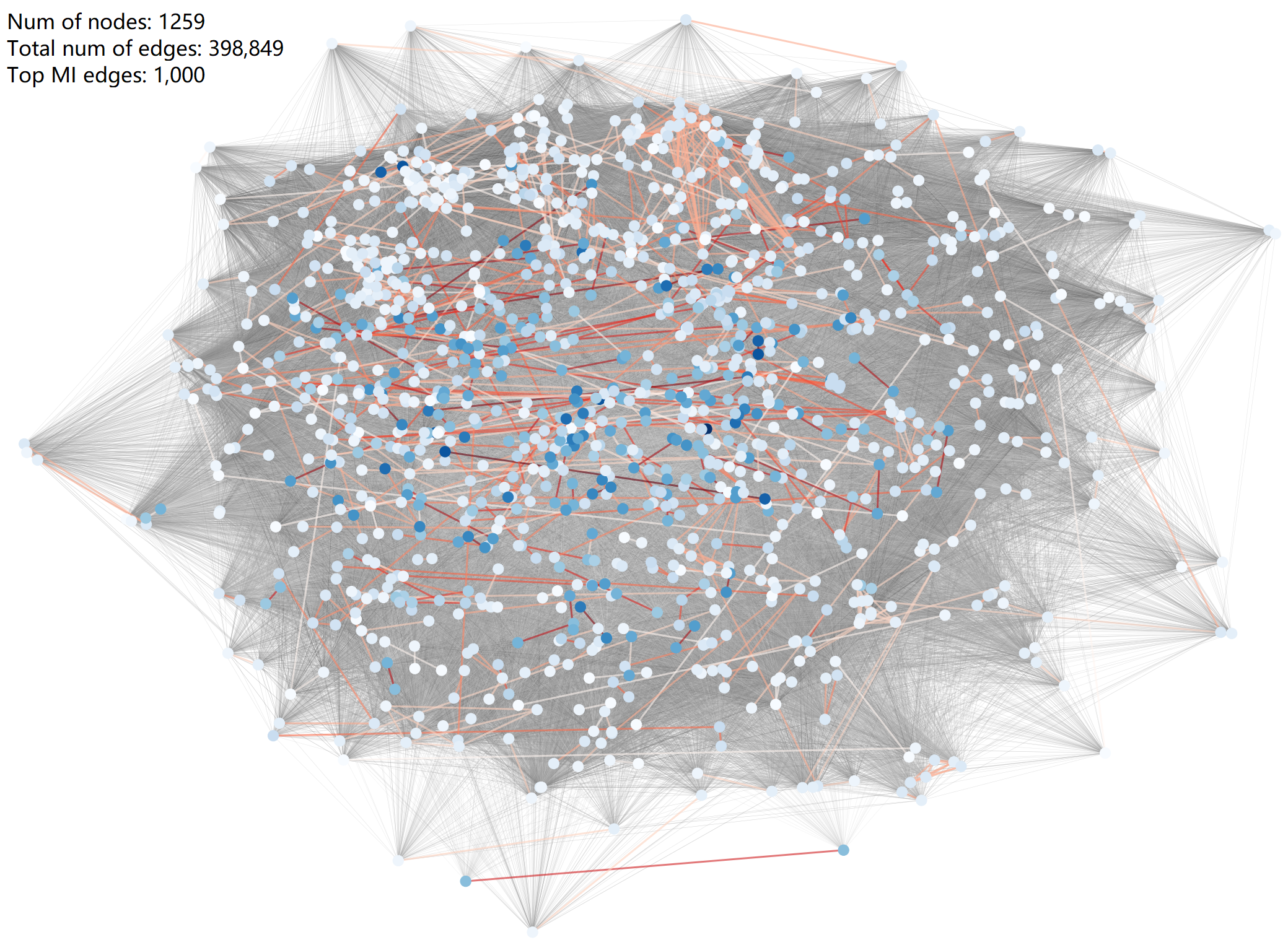}
    \caption{GoT subgraph for the top 1000 MI edges with 1,259 tweet nodes and their edges (top 1000 MI edges and 397,849 other edges). The red edges represent the top 1000 MI edges and the gray edges represent other MI edges. Nodes with higher saturation contains more leading tokens. Similarly, edges with higher saturation indicate larger MI weights.}
    \label{fig:top_mi_got}
    \end{figure}

Subsequently, tweet nodes constituting the top 1000 NMI weights were extracted to construct a new GoT subgraph for further analysis. The 1000 NMI subgraph contained 1,259 tweet nodes, which consisted of 1,024 token nodes. The minimum NMI value within the subgraph was 0.9769. In Figure \ref{fig:top_mi_got}, edges are only drawn between node pairs with intersecting token nodes. The edges with an NMI value of -1 are not shown in the figure. It should be noted that if all the tweet node pairs extracted from the top 1000 MI edges are unique, there would be 2,000 nodes in the subgraph.

\begin{figure}[ht]
    \centering
        \begin{subfigure}{0.235\textwidth}
            \includegraphics[width=\linewidth]{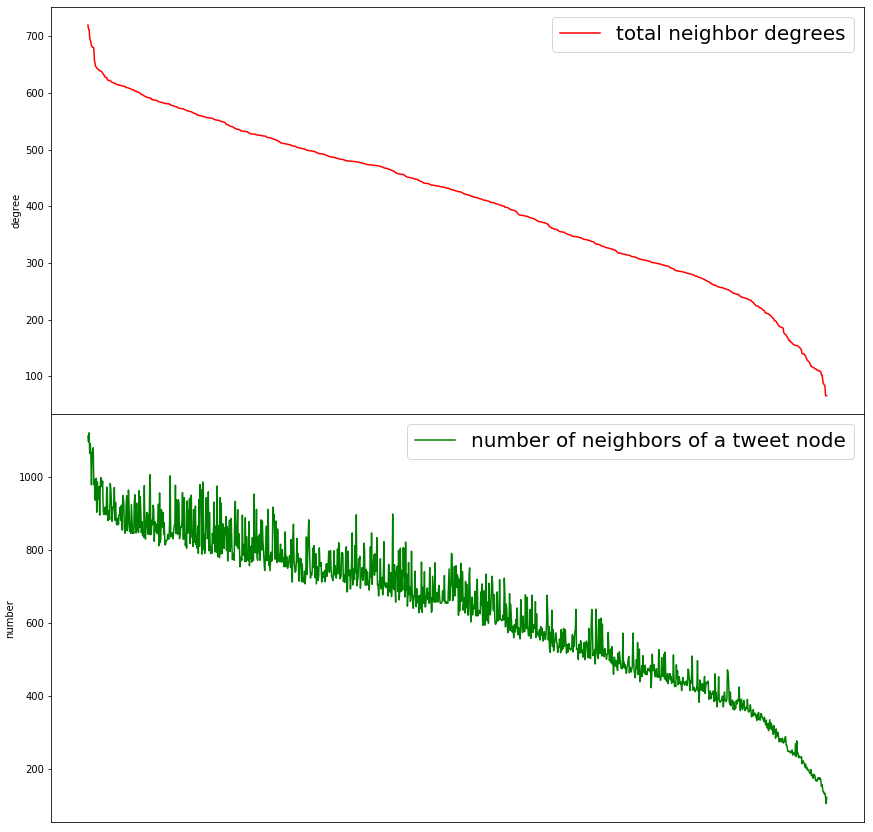}
            \caption{total degrees (top) and number of neighbors (bottom) of tweet nodes ordered by total degrees descendingly.}
            \label{fig:top_mi_degrees_a}
        \end{subfigure} 
        \hfill
        \begin{subfigure}{0.235\textwidth}
        \includegraphics[width=\linewidth]{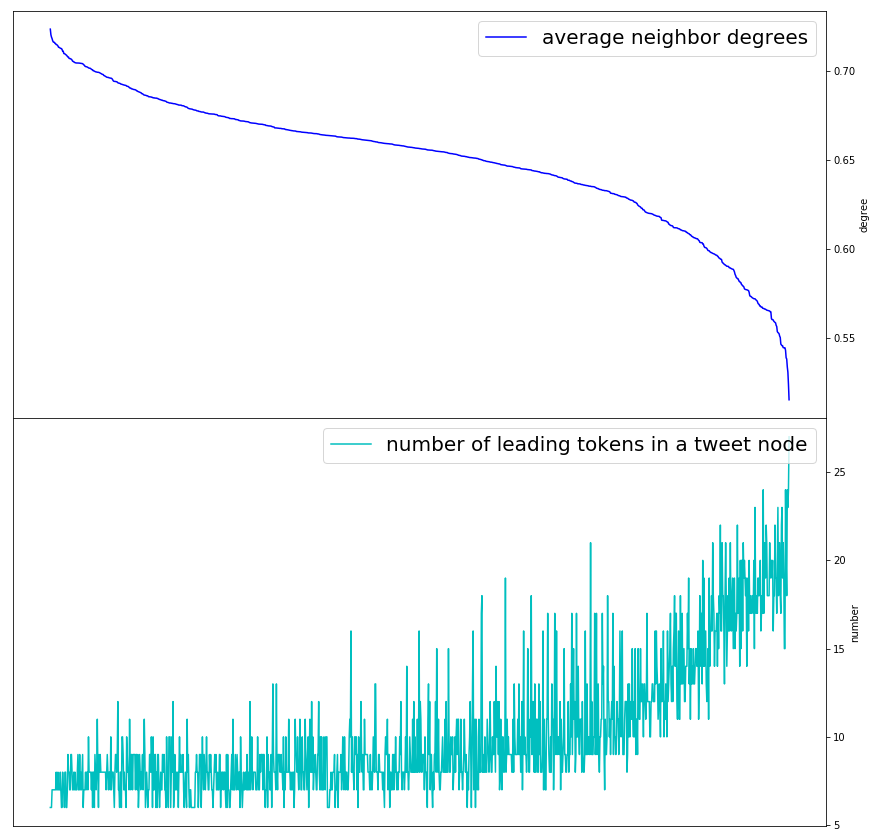}
        \caption{average neighbor degrees (top) and number of leading tokens of tweet nodes ordered by average degrees descendingly.}
        \label{fig:top_mi_degrees_b}
        \end{subfigure}
        \caption{Node degree distribution for top 1000 MI subgraph.}
        \label{fig:top_mi_degrees}
    \end{figure}

After examining the tweet nodes with top 10 total degrees (the sum of all MI weights between the node and its neighbors), we found that some of the nodes are subsets of each other. For instance, \textit{tweet\_23146} is represented as: \{\textit{"news", "epidemiologist", "nation", "probably", "confirmed", "know", "spreading", "humping", "\#mopolitics", "center", "kudla", "called", "contagious", "\#sarscov2", "\#health", "govt", "friday", "\#prevention", "update", "expects", "rrb"}\}, of which two other tweets \textit{tweet\_20023} and \textit{tweet\_11475} are subsets with 2 and 1 token node(s) differences. Further statistics on the associations between total node degrees, average node degrees, and node size (number of leading tokens) are shown in Figure \ref{fig:top_mi_degrees}. 
It can be seen in Figure \ref{fig:top_mi_degrees_a} that total node degree is positively correlated with the number of neighbors. We also examined the correlation between total node degree and average neighbor degree (average NMI weights of all neighbors), but found no correlation. On the other hand, in Figure \ref{fig:top_mi_degrees_b}, the average neighbor degree is negatively correlated with the number of leading tokens in a tweet node. This indicates that NMI relatively favors tweets with less elements. Imagine a random case where only independent tokens are present in a tweet node. Larger nodes (have more tokens) have less chances to share information with other nodes. More precisely, for two pairs of tweets that both share the same intersection size, the size of the tweets will determine the NMI values. For instance, $t_1$ (5 elements) and $t_2$ (5 elements) share 3 elements, while $t_1$ and $t_3$ (10 elements) also share 3 elements. Assuming there are a total of $m=100$ elements, the NMI values as defined in Algorithm \ref{alg:nmi} are $NMI_{t_1, t_2}=0.829$ and $NMI_{t_1, t_3}=0.598$. It should be noted that different normalization methods can cause the NMI values to vary. The normalization metric we chose largely normalizes the upper bound of the MI values. When a different normalization metric is applied, for example, using the joint probability, both the upper and lower bounds of the MI values can be normalized. 

\begin{figure} [ht]
    \centering
    \includegraphics[width=\linewidth]{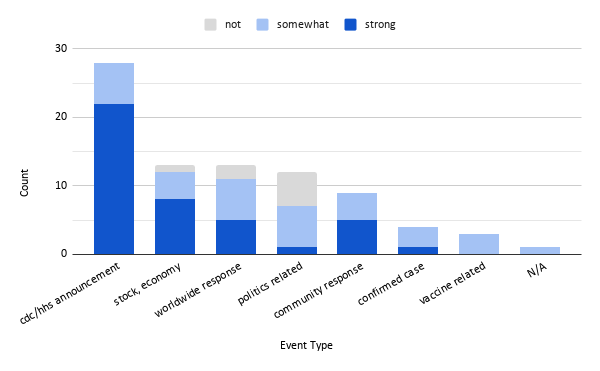}
    \caption{Clique validity distribution by manually labeled event type categories. Deep blue, light blue, and grey correspond to "strong representative", "somewhat representative", and "not representative" categories respectively.}
    \label{fig:clique_result_figure}
\end{figure}

Finally, we identified 83 maximal cliques of size greater than 2 from the top 1000 NMI subgraph. While the largest clique contained 14 tweet nodes, there were 21 cliques consisting of 4 to 6 tweet nodes, and the rests contained 3 tweet nodes. We observed that the tweet nodes contained in the same clique shared highly similar content. Table \ref{tab:clique_example} illustrates the shared token nodes in all 14 tweet nodes from the largest clique. We can derive the information contained in this clique as "the stock and/or bitcoin market has a crash/slide". 
    \begin{table}[ht]
        \caption{Token Nodes Shared Between Tweet Nodes in the Extracted Cliques}
        \centering
        \begin{tabular}{|p{0.4\linewidth}|p{0.5\linewidth}|} \hline
            \textbf{Shared token nodes in all 14 tweet nodes of a strong representative clique} & $\#bitcoinprice$, $\#stockmarket$, $best$, $crashing$, $gdx$, $history$, $probably$, $slide$\\ \hline 
            \textbf{shared token nodes in all 4 tweet nodes of a somewhat representative clique} & $probably$, $hhs$, $open$, $immediate$, $likelihood$, $know$,	$american$ \\ \hline 
        \end{tabular}
        \label{tab:clique_example}
    \end{table}
Further investigation indicated that nodes in this clique are associated with the same tweet: "The Dow just logged its worst 2-day point slide in history — here are 5 reasons the stock market is tanking, and only one of them is the coronavirus", which elaborates on the stock market dropping. We marked this type of clique to be "strong representative" of the sub-event. However, not all cliques represented the original tweet contents well. Another clique in Table \ref{tab:clique_example} that contained the tweets regarding "US HHS says there’s ‘low immediate risk’ of coronavirus for Americans" only suggested the "likelihood of" the main "COVID-19" event. We marked this type of clique as "somewhat representative" of the sub-event. Other types of cliques are marked as "not representative". Table \ref{tab:clique_result} shows the content validity evaluation on 83 maximal cliques. It should be noted that the labels are annotated in acknowledgement of the meaningful token nodes, which on average takes 64.3\% of each clique. 
We also analyzed the event categories among the generated cliques and manually annotated the clique with type labels. Figure \ref{fig:clique_result_figure} shows the content validity distribution by different event types. We can see that our proposed model performed very well on generating meaningful clique content on the categories "health department announcement" and "stock, economy"; and not very well on "politics related" category. The difference in performance may be due to the variation of the amount of information contained in the described events.  

\begin{table}[ht]
    \centering
    \caption{Content Validation of 83 Maximal Cliques}
    \begin{tabular}{|p{0.27\linewidth}|p{0.27\linewidth}|p{0.27\linewidth}|} \hline 
         strong & somewhat & not \\          
         representative & representative & representative \\ \hline
         42 (50.60\%) & 33 (39.76\%)  & 8 (9.64\%) \\ \hline 
    \end{tabular}
    \label{tab:clique_result}
\end{table}

It should be addressed that the dataset used in this paper was collected directly from Twitter and that the approach proposed in this paper was fully automated and unsupervised. As we mentioned previously there is no bench-marking dataset for event identification, and the absence of gold standard makes it difficult to conduct quantitative evaluations on the results. Our best effort, as presented in Table \ref{tab:clique_result} and Figure \ref{fig:clique_result_figure}, was to manually analyzed the content of the generated cliques for validation.

\section{Conclusion and Future Work}
\label{sec:future_work}
In this paper, we proposed a hybrid graph model which uses conceptualized GoW nodes to represent tweet nodes for sub-events identification. We developed an incremental graph merging approach to condense raw GoW leveraging word embeddings. In addition, we outlined how the reduced GoW is connected to GoT and developed an adjusted NMI metric to measure nodes similarity in GoT. Finally, we utilized the fundamental graph structure, cliques, to assist with identifying sub-events. Our approach showed promising results on identifying popular sub-events in a fully unsupervised manner on real-world data. There remain adjustments that can be made to the GoT to improve the robustness of the model so more detailed and less noisy events can be captured. In the future, we plan to employ named entity recognition in raw data processing to identify key concepts. Additionally, frequent bi-grams/n-grams will be examined and combined prior to FastText training to improve the similarity metrics from word embeddings. We will also compare different MI normalization methods to neutralize the bias from sequence length. Finally, we plan to improve the conceptualization method of the token nodes so instead of the leading token, a node can be represented by the concept it is associated to. 

\bibliographystyle{IEEEtran}
\bibliography{WI}

\begin{thebibliography}{10}
\providecommand{\url}[1]{#1}
\csname url@samestyle\endcsname
\providecommand{\newblock}{\relax}
\providecommand{\bibinfo}[2]{#2}
\providecommand{\BIBentrySTDinterwordspacing}{\spaceskip=0pt\relax}
\providecommand{\BIBentryALTinterwordstretchfactor}{4}
\providecommand{\BIBentryALTinterwordspacing}{\spaceskip=\fontdimen2\font plus
\BIBentryALTinterwordstretchfactor\fontdimen3\font minus
  \fontdimen4\font\relax}
\providecommand{\BIBforeignlanguage}[2]{{%
\expandafter\ifx\csname l@#1\endcsname\relax
\typeout{** WARNING: IEEEtran.bst: No hyphenation pattern has been}%
\typeout{** loaded for the language `#1'. Using the pattern for}%
\typeout{** the default language instead.}%
\else
\language=\csname l@#1\endcsname
\fi
#2}}
\providecommand{\BIBdecl}{\relax}
\BIBdecl

\bibitem{feng:2015}
W.~Feng, C.~Zhang, W.~Zhang, J.~Han, J.~Wang, C.~Aggarwal, and J.~Huang,
  ``Streamcube: hierarchical spatio-temporal hashtag clustering for event
  exploration over the twitter stream,'' in \emph{2015 IEEE 31st International
  Conference on Data Engineering}.\hskip 1em plus 0.5em minus 0.4em\relax IEEE,
  2015, pp. 1561--1572.

\bibitem{yang:2018}
S.-F. Yang and J.~T. Rayz, ``An event detection approach based on twitter
  hashtags,'' \emph{arXiv preprint arXiv:1804.11243}, 2018.

\bibitem{mcminn:2015}
A.~J. McMinn and J.~M. Jose, ``Real-time entity-based event detection for
  twitter,'' in \emph{International conference of the cross-language evaluation
  forum for european languages}.\hskip 1em plus 0.5em minus 0.4em\relax
  Springer, Cham, 2015, pp. 65--77.

\bibitem{qin:2018}
Y.~Qin, Y.~Zhang, M.~Zhang, and D.~Zheng, ``Frame-based representation for
  event detection on twitter,'' \emph{IEICE TRANSACTIONS on Information and
  Systems}, vol. 101, no.~4, pp. 1180--1188, 2018.

\bibitem{dhingra:2016}
B.~Dhingra, Z.~Zhou, D.~Fitzpatrick, M.~Muehl, and W.~W. Cohen, ``Tweet2vec:
  Character-based distributed representations for social media,'' \emph{arXiv
  preprint arXiv:1605.03481}, 2016.

\bibitem{pina:2016}
L.~N. Pina and R.~Johansson, ``Embedding senses for efficient graph-based word
  sense disambiguation,'' in \emph{Proceedings of TextGraphs-10: the workshop
  on graph-based methods for natural language processing}, 2016, pp. 1--5.

\bibitem{bevilacqua:2020}
M.~Bevilacqua and R.~Navigli, ``Breaking through the 80\% glass ceiling:
  Raising the state of the art in word sense disambiguation by incorporating
  knowledge graph information,'' in \emph{Proceedings of the 58th Annual
  Meeting of the Association for Computational Linguistics}, 2020, pp.
  2854--2864.

\bibitem{skianis:2018}
K.~Skianis, F.~Malliaros, and M.~Vazirgiannis, ``Fusing document, collection
  and label graph-based representations with word embeddings for text
  classification,'' in \emph{Proceedings of the Twelfth Workshop on Graph-Based
  Methods for Natural Language Processing ({T}ext{G}raphs-12)}.\hskip 1em plus
  0.5em minus 0.4em\relax ACL, 2018, pp. 49--58.

\bibitem{yao:2019}
L.~Yao, C.~Mao, and Y.~Luo., ``Graph convolutional networks for text
  classification,'' in \emph{Proceedings of the AAAI Conference on Artificial
  Intelligence}, vol.~33, 2019, pp. 7370--7377.

\bibitem{nayeem:2017}
M.~T. Nayeem and Y.~Chali, ``Extract with order for coherent multi-document
  summarization,'' in \emph{Proceedings of {T}ext{G}raphs-11: the Workshop on
  Graph-based Methods for Natural Language Processing}.\hskip 1em plus 0.5em
  minus 0.4em\relax Vancouver, Canada: ACL, 2017, pp. 51--56.

\bibitem{yasunaga:2017}
M.~Yasunaga, R.~Zhang, K.~Meelu, A.~Pareek, K.~Srinivasan, and D.~Radev,
  ``Graph-based neural multi-document summarization,'' \emph{arXiv preprint
  arXiv:1706.06681}, 2017.

\bibitem{tonon:2017}
A.~Tonon, P.~Cudr{\'e}-Mauroux, A.~Blarer, V.~Lenders, and B.~Motik,
  ``Armatweet: detecting events by semantic tweet analysis,'' in \emph{European
  Semantic Web Conference}.\hskip 1em plus 0.5em minus 0.4em\relax Springer,
  2017, pp. 138--153.

\bibitem{fedoryszak:2019}
M.~Fedoryszak, B.~Frederick, V.~Rajaram, and C.~Zhong, ``Real-time event
  detection on social data streams,'' in \emph{Proceedings of the 25th ACM
  SIGKDD International Conference on Knowledge Discovery \& Data Mining}, 2019,
  pp. 2774--2782.

\bibitem{vazirgiannis:2018}
M.~Vazirgiannis, F.~D. Malliaros, and G.~Nikolentzos., ``Graphrep: boosting
  text mining, nlp and information retrieval with graphs,'' in
  \emph{Proceedings of the 27th ACM International Conference on Information and
  Knowledge Management}, 2018, pp. 2295--2296.

\bibitem{miller:1995}
G.~A. Miller, ``Wordnet: a lexical database for english,'' \emph{Communications
  of the ACM}, vol.~38, no.~11, pp. 39--41, 1995.

\bibitem{bollacker:2008}
K.~Bollacker, C.~Evans, P.~Paritosh, T.~Sturge, and J.~Taylor, ``Freebase: a
  collaboratively created graph database for structuring human knowledge,'' in
  \emph{Proceedings of the 2008 ACM SIGMOD international conference on
  Management of data}, 2008, pp. 1247--1250.

\bibitem{jin:2016}
C.-X. Jin and Q.-C. Bai, ``Text clustering algorithm based on the graph
  structures of semantic word co-occurrence,'' in \emph{2016 International
  Conference on Information System and Artificial Intelligence (ISAI)}.\hskip
  1em plus 0.5em minus 0.4em\relax IEEE, 2016, pp. 497--502.

\bibitem{jinarat:2018}
S.~Jinarat, B.~Manaskasemsak, and A.~Rungsawang, ``Short text clustering based
  on word semantic graph with word embedding model.'' in \emph{2018 Joint 10th
  International Conference on Soft Computing and Intelligent Systems (SCIS) and
  19th International Symposium on Advanced Intelligent Systems (ISIS)}.\hskip
  1em plus 0.5em minus 0.4em\relax IEEE, 2018, pp. 1427--1432.

\bibitem{word2vec:2013}
T.~Mikolov, I.~Sutskever, K.~Chen, G.~S. Corrado, and J.~Dean, ``Distributed
  representations of words and phrases and their compositionality,'' in
  \emph{Advances in neural information processing systems}, 2013.

\bibitem{meladianos:2015}
P.~Meladianos, G.~Nikolentzos, F.~Rousseau, Y.~Stavrakas, and M.~Vazirgiannis.,
  ``Degeneracy-based real-time sub-event detection in twitter stream,'' in
  \emph{Ninth international AAAI conference on web and social media}, 2015.

\bibitem{parveen:2014}
D.~Parveen and M.~Strube, ``Multi-document summarization using bipartite
  graphs,'' in \emph{Proceedings of {T}ext{G}raphs-9: the workshop on
  Graph-based Methods for Natural Language Processing}.\hskip 1em plus 0.5em
  minus 0.4em\relax Doha, Qatar: ACL, 2014, pp. 15--24.

\bibitem{hits:1999}
J.~M. Kleinberg, ``Authoritative sources in a hyperlinked environment,''
  \emph{Journal of the ACM (JACM)}, vol.~46, no.~5, pp. 604--632, 1999.

\bibitem{mihalcea:2004}
R.~Mihalcea and P.~Tarau, ``Textrank: Bringing order into text,'' in
  \emph{Proceedings of the 2004 conference on empirical methods in natural
  language processing}, 2004, pp. 404--411.

\bibitem{glavavs:2016}
G.~Glava{\v{s}}, F.~Nanni, and S.~P. Ponzetto, ``Unsupervised text segmentation
  using semantic relatedness graphs.''\hskip 1em plus 0.5em minus 0.4em\relax
  Association for Computational Linguistics, 2016.

\bibitem{szubert:2019}
I.~Szubert and M.~Steedman, ``Node embeddings for graph merging: Case of
  knowledge graph construction,'' in \emph{Proceedings of the Thirteenth
  Workshop on Graph-Based Methods for Natural Language Processing
  (TextGraphs-13)}, 2019, pp. 172--176.

\bibitem{harrington:2008}
B.~Harrington and S.~Clark, ``Asknet: Creating and evaluating large scale
  integrated semantic networks,'' \emph{International Journal of Semantic
  Computing}, vol.~2, no.~03, pp. 343--364, 2008.

\bibitem{lin:2004}
C.-Y. Lin, ``Rouge: A package for automatic evaluation of summaries,'' in
  \emph{Text summarization branches out}, 2004, pp. 74--81.

\bibitem{firth:1957}
J.~R. Firth, ``A synopsis of linguistic theory,'' \emph{Studies in linguistic
  analysis.}, pp. 1930--1955, 1957.

\bibitem{fasttext:2017}
P.~Bojanowski, E.~Grave, A.~Joulin, and T.~Mikolov, ``Enriching word vectors
  with subword information,'' \emph{Transactions of the Association for
  Computational Linguistics}, vol.~5, pp. 35--146., 2017.

\bibitem{church:1990}
K.~Church and P.~Hanks, ``Word association norms, mutual information, and
  lexicography,'' \emph{Computational linguistics}, vol.~16, no.~1, pp. 22--29,
  1990.

\bibitem{bouma:2009}
G.~Bouma, ``Normalized (pointwise) mutual information in collocation
  extraction,'' in \emph{Proceedings of GSCL}, 2009, pp. 31--40.

\bibitem{amelio:2017}
A.~Amelio and C.~Pizzuti, ``Correction for closeness: Adjusting normalized
  mutual information measure for clustering comparison,'' \emph{Computational
  Intelligence}, vol.~33, no.~3, pp. 579--601, 2017.

\bibitem{romano:2014}
S.~Romano, J.~Bailey, V.~Nguyen, and K.~Verspoor, ``Standardized mutual
  information for clustering comparisons: one step further in adjustment for
  chance.'' in \emph{International Conference on Machine Learning}, 2014, pp.
  1143--1151.

\end{thebibliography}

\end{document}